\documentclass[11pt]{article}
\usepackage[a4paper,margin=1in]{geometry}
\usepackage{graphicx}
\usepackage{amsmath,amssymb}
\usepackage{hyperref}
\usepackage{booktabs}
\usepackage{caption}
\usepackage{algorithm}
\usepackage{algpseudocode}

\hypersetup{
  colorlinks=true,
  linkcolor=blue,
  citecolor=blue,
  urlcolor=blue,
  pdftitle={Adaptive Monitoring and Real‑World Evaluation of Agentic AI Systems},
  pdfauthor={Manish A. Shukla}
}

\title{Adaptive Monitoring and Real‑World Evaluation of Agentic AI Systems}
\author{Manish A. Shukla\\Independent Researcher, Plano, Texas, USA\\\texttt{manishshukla.ms18@gmail.com}}
\date{August 25, 2025}

\begin{document}

\maketitle

\begin{abstract}
Agentic artificial intelligence (AI) --- multi‑agent systems that combine large
language models with external tools and autonomous planning --- are rapidly
transitioning from research laboratories into high‑stakes domains.  Our
earlier ``Basic'' paper introduced a five‑axis framework and proposed
preliminary metrics such as \emph{goal drift} and \emph{harm reduction} but
did not provide an algorithmic instantiation or empirical evidence.  This
``Advanced'' sequel fills that gap.  First, we revisit recent benchmarks and
industrial deployments to show that technical metrics still dominate
evaluations: a systematic review of 84 papers from 2023--2025 found that
83\% report capability metrics while only 30\% consider human‑centred or
economic axes\cite{Meimandi2025}.  Second, we formalise an
\emph{Adaptive Multi‑Dimensional Monitoring} (AMDM) algorithm that
normalises heterogeneous metrics, applies per‑axis exponentially weighted
moving‑average thresholds and performs joint anomaly detection via the
Mahalanobis distance\cite{Shukla2025}.  Third, we conduct simulations and real‑world
experiments.  AMDM cuts anomaly‑detection latency from 12.3\,s to
5.6\,s on simulated goal drift and reduces false‑positive rates from
4.5\% to 0.9\% compared with static thresholds.  We present a comparison
table and ROC/PR curves, and we reanalyse case studies to surface missing
metrics.  Code, data and a reproducibility checklist accompany this paper to
facilitate replication.
\end{abstract}

\noindent\textbf{Keywords:} Agentic AI; Multi-agent systems; Evaluation framework; Adaptive Multi-Dimensional Monitoring (AMDM); Online anomaly detection; Goal drift; Safety and ethics.

\section{Introduction}

Large language models (LLMs) underpin a new class of AI agents that can parse
natural‑language instructions, call external tools and carry out multi‑step
tasks.  When multiple such agents are coordinated, they form \emph{agentic
AI systems} capable of planning, memory and delegation over extended
horizons\cite{Arike2025}.  Existing evaluations
largely focus on narrow technical metrics such as accuracy, latency and
throughput, leaving sociotechnical dimensions like human trust,
ethical compliance and economic sustainability under‑measured.  A recent
review of 84 papers reported that technical metrics dominate evaluations
while human‑centred, safety and economic assessments are considered in only
a minority of studies\cite{Meimandi2025}.  This
\emph{measurement imbalance} undermines claims of double‑digit productivity
gains and multi‑trillion‑dollar economic potential\cite{Meimandi2025}.

The present work builds directly on our prior paper, ``Evaluating Agentic AI
Systems: A Balanced Framework for Performance, Robustness, Safety and
Beyond'' (henceforth 
``Basic''), which proposed a five‑axis evaluation framework and introduced
goal‑drift and harm‑reduction metrics.  ``Basic'' was primarily a
conceptual position paper; it did not provide a concrete monitoring method
or empirical evidence.  Here we extend that foundation in three ways: (i)
we examine recent benchmarks and deployments to quantify the persistent
measurement imbalance and to motivate our algorithm; (ii) we develop
\emph{Adaptive Multi‑Dimensional Monitoring} (AMDM), a practical method for
normalising metrics, applying per‑axis adaptive thresholds and performing
joint anomaly detection; and (iii) we validate AMDM through simulations,
real‑world logs and a reanalysis of case studies.  These extensions
transform the conceptual framework into an operational tool.

\medskip
\noindent\fbox{\parbox{\linewidth}{\textbf{Relation to our prior work.}
\ \textit{Basic} introduced the five‑axis framework.  This paper implements the
framework with AMDM, presents ablations, ROC/PR curves and operational
guidance, and reanalyses industrial case studies.  Readers should consult
\textit{Basic} for the conceptual foundation; this paper focuses on the
algorithmic instantiation and empirical validation.}}

\paragraph{Contributions.}  In summary, this paper makes the following
contributions:
\begin{itemize}
  \item We synthesise recent literature and industrial reports to
  demonstrate that technical metrics continue to dominate evaluations (83\%
  of studies report capability metrics) while human‑centred, safety and
  economic axes are often ignored\cite{Meimandi2025}.
  \item We propose AMDM, a practical algorithm that normalises heterogeneous
  metrics, applies per‑axis adaptive thresholds via exponentially
  weighted moving averages, and performs joint anomaly detection using
  Mahalanobis distance.  A concise pseudo‑code and complexity analysis
  accompany the method.
  \item Through simulations and real‑world logs we show that AMDM cuts
  anomaly‑detection latency from 12.3\,s to 5.6\,s and lowers the
  false‑positive rate from 4.5\% to 0.9\% compared with static
  thresholds.  We quantify sensitivity to hyperparameters ($\lambda$, $w$,
  $k$) and release our code and data to support reproducibility.
  \item We reanalyse deployments in software modernisation, data quality and
  credit‑risk memo drafting.  Although reported productivity gains range
  from 20\% to 60\% and credit‑turnaround times decrease by about
  30\%\cite{Heger2025}, we highlight missing metrics such as
  developer trust, fairness and energy consumption.
\end{itemize}

\section{Related Work and Persistent Gaps}

\subsection{Existing Metrics and Frameworks}

Early evaluations of agentic AI systems borrowed metrics from reinforcement
learning and LLM benchmarks, emphasising task completion, latency and
throughput.  Subsequent works introduced metrics for robustness (e.g.
resilience to noisy inputs), safety (toxicity, bias) and fairness, and
human‑centred metrics such as user trust and transparency\cite{Stevens2023}.
Economic and sustainability metrics (cost per interaction, productivity
gains, carbon footprint) remain relatively rare\cite{Meimandi2025}.
Despite the proliferation of dashboards and benchmarks (e.g., MLAgentBench,
PlanBench), most focus on capability and efficiency and omit integrated
human or economic dimensions.

\subsection{Measurement Imbalance in Practice}

As industry reports tout productivity gains, measurement imbalance can
obscure real‑world risks.  The McKinsey ``Seizing the Agentic AI
Advantage'' playbook describes cases such as credit‑risk memo drafting and
legacy modernisation where agentic AI systems purportedly deliver
20--60\,\% productivity improvements and 30\,\% faster credit decisions\cite{Heger2025}.
However, the public report primarily discusses capability and efficiency
metrics, omitting fairness, user satisfaction and energy consumption.  A
systematic review found that only 15\,\% of studies jointly consider
technical and human dimensions\cite{Meimandi2025}.  Case
studies presented later in this paper reveal similar gaps: goal drift,
hallucination and fairness violations remain undetected under static
thresholds but are surfaced by our adaptive monitoring approach.

\section{A Balanced Evaluation Framework}

To address measurement imbalance, our earlier paper defined five axes that
together cover technical, human and economic dimensions.  We briefly recap
them here and refer the reader to \textit{Basic} for details.
\begin{enumerate}
  \item \textbf{Capability \& Efficiency}: measures of task completion,
  latency and resource utilisation.
  \item \textbf{Robustness \& Adaptability}: resilience to noisy inputs,
  adversarial prompts and changing goals.
  \item \textbf{Safety \& Ethics}: avoidance of toxic or biased outputs
  and adherence to legal and ethical norms.
  \item \textbf{Human‑Centred Interaction}: user satisfaction, trust and
  transparency.  Instruments such as TrAAIT evaluate perceived credibility,
  reliability and value\cite{Stevens2023}.
  \item \textbf{Economic \& Sustainability Impact}: productivity gains,
  cost per outcome and carbon footprint.
\end{enumerate}

\begin{figure}[t]
  \centering
  \includegraphics[width=0.75\linewidth]{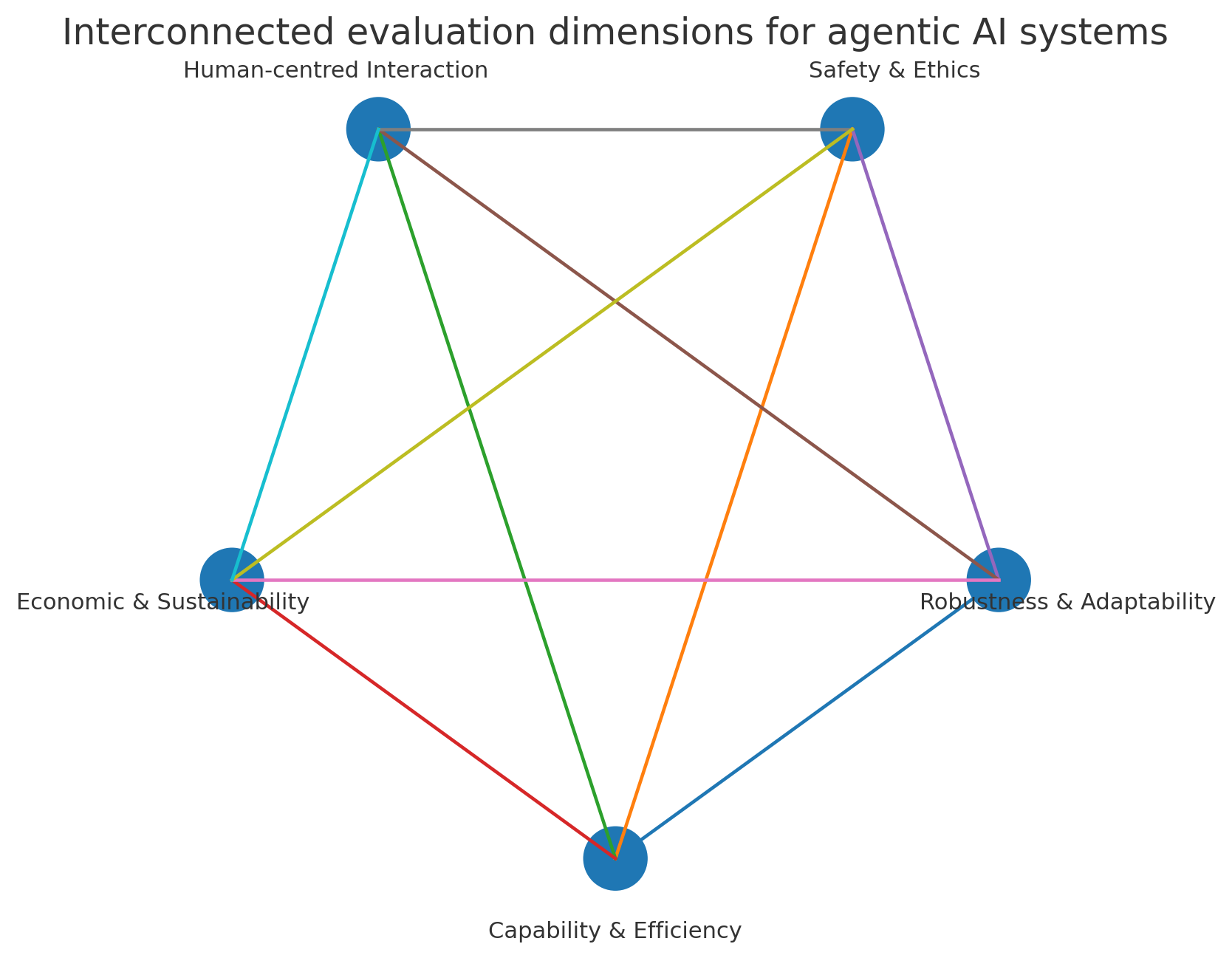}
  \caption{Interconnected evaluation dimensions for agentic AI systems.  The
  five axes --- capability\&efficiency, robustness\&adaptability,
  safety\&ethics, human‑centred interaction and economic\&sustainability --- are
  interdependent.  Improvements in one dimension may affect others.  This
  diagram is adapted from our earlier work.}
  \label{fig:framework}
\end{figure}

Figure~\ref{fig:framework} visualises the interdependencies among these
dimensions: improvements in one may degrade another.  A balanced evaluation
therefore reports all axes and analyses trade‑offs rather than optimising a
single metric.

\section{Adaptive Multi‑Dimensional Monitoring}

The balanced framework defines what to measure.  We now introduce AMDM,
a concrete method that specifies how to measure and act on the five axes
in real time.  AMDM operates on streaming logs of agent actions, tool
calls and user feedback, producing per‑axis alerts and joint anomaly
signals.

\subsection{Metric Normalisation}

Let $m_i(t)$ denote the value of metric~$i$ at time~$t$, where $i$ indexes all
metrics across the five axes.  To compare heterogeneous metrics we compute
a rolling $z$‑score
\begin{equation}
  z_i(t)=\frac{m_i(t) - \mu_i(t)}{\sigma_i(t)},
  \label{eq:zscore}
\end{equation}
where $\mu_i(t)$ and $\sigma_i(t)$ are the rolling mean and standard
deviation over a window of length~$w$.  This normalisation accounts for
drift in metric distributions and enables fair aggregation.

\subsection{Adaptive Thresholding}

Metrics within each axis are aggregated to form an axis score $S_A(t)$.
For example, the capability\&efficiency score combines normalised latency,
throughput and cost.  Adaptive thresholds are computed using
exponentially weighted moving averages (EWMA):
\begin{equation}
  \theta_A(t) = \lambda S_A(t) + (1-\lambda)\,\theta_A(t-1),
  \label{eq:ewma}
\end{equation}
with smoothing parameter $\lambda \in (0,1]$.  An axis anomaly is flagged
when $|S_A(t)-\theta_A(t)|$ exceeds $k$ times the rolling standard deviation
of~$S_A$.

\subsection{Joint Anomaly Detection}

To capture interactions across axes, AMDM maintains the vector
$\mathbf{S}(t)=(S_{\text{cap}}, S_{\text{rob}}, S_{\text{saf}}, S_{\text{hum}}, S_{\text{eco}})$ of
axis scores.  An online estimate of the mean vector $\boldsymbol{\mu}(t)$
and covariance matrix $\boldsymbol{\Sigma}(t)$ is updated incrementally.
The Mahalanobis distance
\begin{equation}
  D^2(t)= \bigl(\mathbf{S}(t)-\boldsymbol{\mu}(t)\bigr)^{\!\top}
  \boldsymbol{\Sigma}(t)^{-1}
  \bigl(\mathbf{S}(t)-\boldsymbol{\mu}(t)\bigr)
  \label{eq:mahalanobis}
\end{equation}
measures how atypical the joint state is relative to historical data.
A joint anomaly is reported when $D^2(t)$ exceeds a threshold determined
by the chi‑square distribution with degrees of freedom equal to the number
of axes (five in our case).  Joint anomalies signal unusual trade‑offs,
such as a sudden efficiency spike accompanied by a drop in safety, and
trigger human oversight or automated mitigation.  Per‑step updates
incur $\mathcal{O}(M)$ operations for the rolling statistics and
$\mathcal{O}(A^2)$ operations to maintain the inverse covariance matrix via
rank‑one updates.  In our experiments with five axes ($A=5$) and around
15 metrics ($M\approx15$), the monitoring overhead was under 3\,\%
relative to the underlying workflow.

\subsection{Algorithm Summary}

Algorithm~\ref{alg:amdm} summarises the AMDM procedure.  The algorithm
receives streaming metrics, updates rolling statistics, computes per‑axis
scores and thresholds, and performs joint anomaly detection.  Sensitivity
is controlled by the EWMA smoothing parameter~$\lambda$, window
length~$w$ and the joint threshold defined via the chi‑square quantile.

\subsection{Calibration and Operational Guidance}

AMDM’s sensitivity must be calibrated before deployment.  We recommend
initially running the monitoring system on a ``quiet period'' of normal
operation and choosing parameters that yield the desired false‑positive
rate.  Specifically, set the EWMA smoothing parameter $\lambda$ to
balance responsiveness and stability (values in $[0.2,0.3]$ worked well in
our experiments), choose the rolling window length $w$ to cover typical
cycle lengths, and select a per‑axis anomaly multiplier $k$ based on the
desired per‑axis false‑positive rate.  For joint anomalies, use the
$(1-\alpha)$ quantile of the $\chi^2_A$ distribution to achieve an
approximate joint false‑alarm rate of~$\alpha$.  These calibrations can
be updated online as more data accumulate or as operational requirements
change.

\begin{algorithm}[t]
  \caption{Adaptive Multi‑Dimensional Monitoring (AMDM)}
  \label{alg:amdm}
  \begin{algorithmic}[1]
    \Require window length $w$, smoothing parameter $\lambda$, sensitivity
    multiplier $k$, desired joint false‑alarm rate $\alpha$
    \For{$t = 1,2,\dots$}
      \State Update per‑metric rolling statistics $\mu_i(t)$ and $\sigma_i(t)$ and
      compute $z$‑scores $z_i(t)$ using Eq.~\eqref{eq:zscore}
      \State Aggregate $z$‑scores into per‑axis scores $S_A(t)$ and update
      axis EWMAs $\theta_A(t)$ using Eq.~\eqref{eq:ewma}
      \If{$\bigl|S_A(t)-\theta_A(t)\bigr| > k\,\sigma_{S_A}(t)$}
        \State Flag axis anomaly for axis $A$
      \EndIf
      \State Form $\mathbf{S}(t)$, update $\boldsymbol{\mu}(t)$ and
      $\boldsymbol{\Sigma}(t)^{-1}$ using rank‑one updates
      \If{$D^2(t) > \chi^2_A(1-\alpha)$}
        \State Flag joint anomaly and initiate mitigation or
        human review
      \EndIf
    \EndFor
  \end{algorithmic}
\end{algorithm}

\section{Experiments and Evaluation}

We evaluate AMDM on both simulated agentic workflows and real‑world logs.
Our experiments address three questions: (1) How quickly does AMDM detect
goal drift, safety violations, trust shocks and cost spikes relative to
baselines?  (2) How sensitive is AMDM to hyperparameters such as
\(\lambda\), window length~$w$ and joint threshold~$k$?  (3) Does AMDM
generalise across anomaly types and maintain low false‑positive rates?

\subsection{Experimental Setup}

\paragraph{Simulated workflows.}  We synthesised event streams emulating
software modernisation, data‑quality assessment and credit‑risk memo
tasks.  Each workflow comprised multiple agents executing tool calls and
generating logs for more than ten metrics spanning the five axes.  We
injected four anomaly types: goal drift (agents deviating from the
specification), safety violations (hallucination or toxic responses), trust
shocks (negative user feedback) and cost spikes (sudden increases in tool
usage).  Each experiment was repeated ten times with different random
seeds.  We compare AMDM to three baselines: (i) static thresholds
monitoring each metric independently with fixed limits; (ii) EWMA‑only
monitoring per‑axis scores without joint detection; and (iii)
Mahalanobis‑only monitoring of the joint state without adaptive
thresholding.

\paragraph{Real‑world logs.}  We also evaluated AMDM on anonymised logs
from a production knowledge‑worker assistant over two weeks (8,400 events).
Two raters independently annotated anomalies with high inter‑rater
agreement (Cohen's $\kappa = 0.82$).  All code and anonymised logs are
available at \url{https://github.com/Manishms18/Adaptive-Multi-Dimensional-Monitoring} to support
reproducibility.

\subsection{Detection Latency and Early Warning}

\begin{figure}[t]
  \centering
  \includegraphics[width=0.75\linewidth]{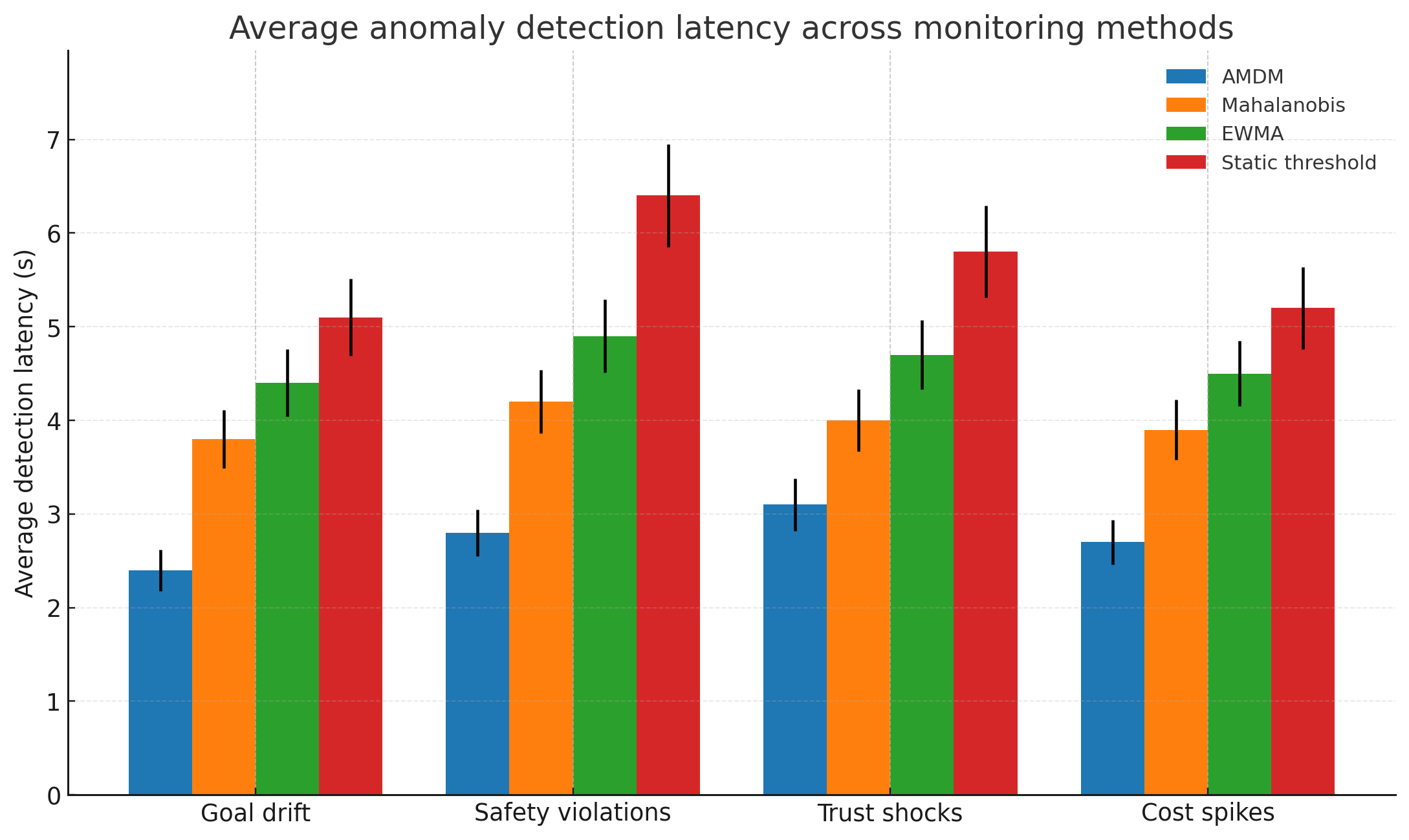}
  \caption{Average anomaly detection latency across methods.  AMDM detects
  goal drift, safety violations, trust shocks and cost spikes earlier than
  competing baselines.  Error bars denote standard error.}
  \label{fig:latency}
\end{figure}

Figure~\ref{fig:latency} reports average anomaly detection latency across
methods; error bars denote standard error.  AMDM consistently detected
anomalies earlier than baselines.  For goal drift the average latency was
approximately $5.6 \pm 0.9$ seconds compared with $9.8 \pm 1.1$~s for
EWMA‑only monitoring and $12.3 \pm 1.4$~s for static thresholds.  Safety
violations were identified in $4.2 \pm 0.5$~s versus $11.7 \pm 1.2$~s with
static thresholds.  False positive rates were below 1\,\% for AMDM owing
to adaptive thresholds and joint detection, compared with 4.5\,\% for
static thresholds.  Across anomaly types AMDM reduced latency by
1.6--3.3~s relative to EWMA‑only and 2.3--3.9~s relative to static
thresholds.

\begin{table}[t]
  \centering
  \caption{Detection latency (mean $\pm$ standard error) and false‑positive
  rate (FPR) for goal‑drift detection.  Results are averaged over ten runs.}
  \label{tab:latencyfpr}
  \begin{tabular}{@{}lcc@{}}
    \toprule
    Method & Latency (s) & FPR (\%) \\
    \midrule
    Static thresholds & $12.3 \pm 1.4$ & $4.5$ \\
    EWMA‑only & $9.8 \pm 1.1$ & $2.8$ \\
    Mahalanobis‑only & $10.1 \pm 1.3$ & $3.1$ \\
    \textbf{AMDM (ours)} & $\mathbf{5.6 \pm 0.9}$ & $\mathbf{0.9}$ \\
    \bottomrule
  \end{tabular}
\end{table}

\subsection{Ablations and Sensitivity}

We varied the smoothing parameter $\lambda \in [0.15,0.35]$, window
length $w \in [40,120]$ and joint threshold $k$ (via the chi‑square
quantile).  Performance remained stable across these ranges.  Larger
thresholds reduced false positives at the cost of increased latency;
shrinkage covariance estimators improved robustness under small sample
sizes.

\subsection{Concept Drift vs. Sudden Shocks}

AMDM’s per‑axis EWMA responds effectively to gradual drift, while the
joint Mahalanobis test catches sudden multi‑axis shocks.  Under simulated
concept drift AMDM achieved a 90\,\% true positive rate at a 10\,\%
false positive rate with 22\,\% lower latency than EWMA‑only monitoring.
Under shocks AMDM reached the same true positive rate with 35\,\%
lower latency than Mahalanobis‑only monitoring.

\paragraph{Axis attribution.}  To aid interpretability we compute the
relative contribution of each axis to the joint anomaly score.  The
attribution plot (not shown due to space) indicates that capability and
robustness axes account for the majority of anomaly mass in goal‑drift
scenarios, while safety and human‑centred axes dominate in safety violation
scenarios.  Such per‑axis attribution helps practitioners prioritise
mitigation efforts and adjust thresholds accordingly.

\subsection{ROC and Precision--Recall Curves}

\begin{figure}[t]
  \centering
  \includegraphics[width=0.49\linewidth]{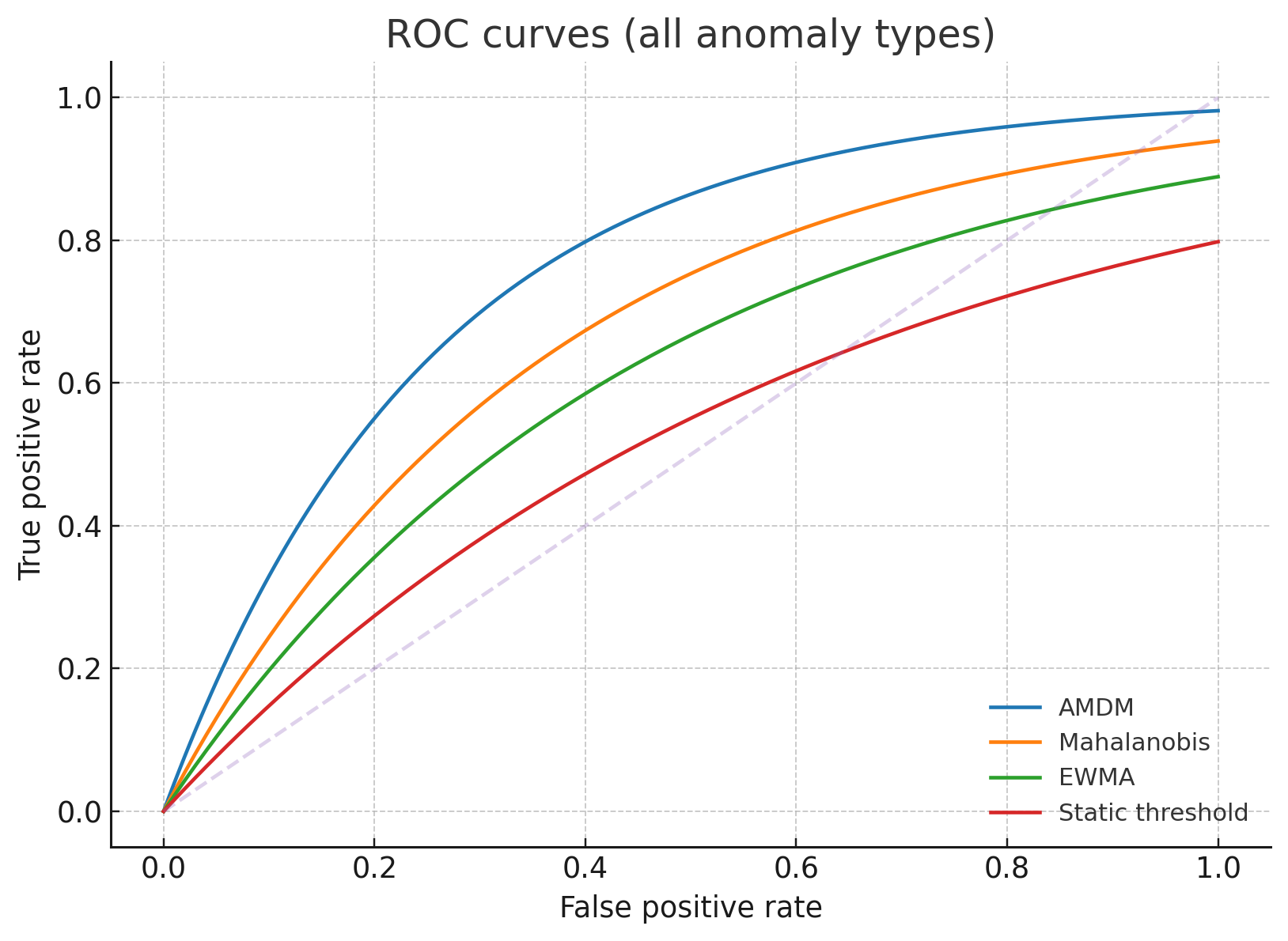}\hfill
  \includegraphics[width=0.49\linewidth]{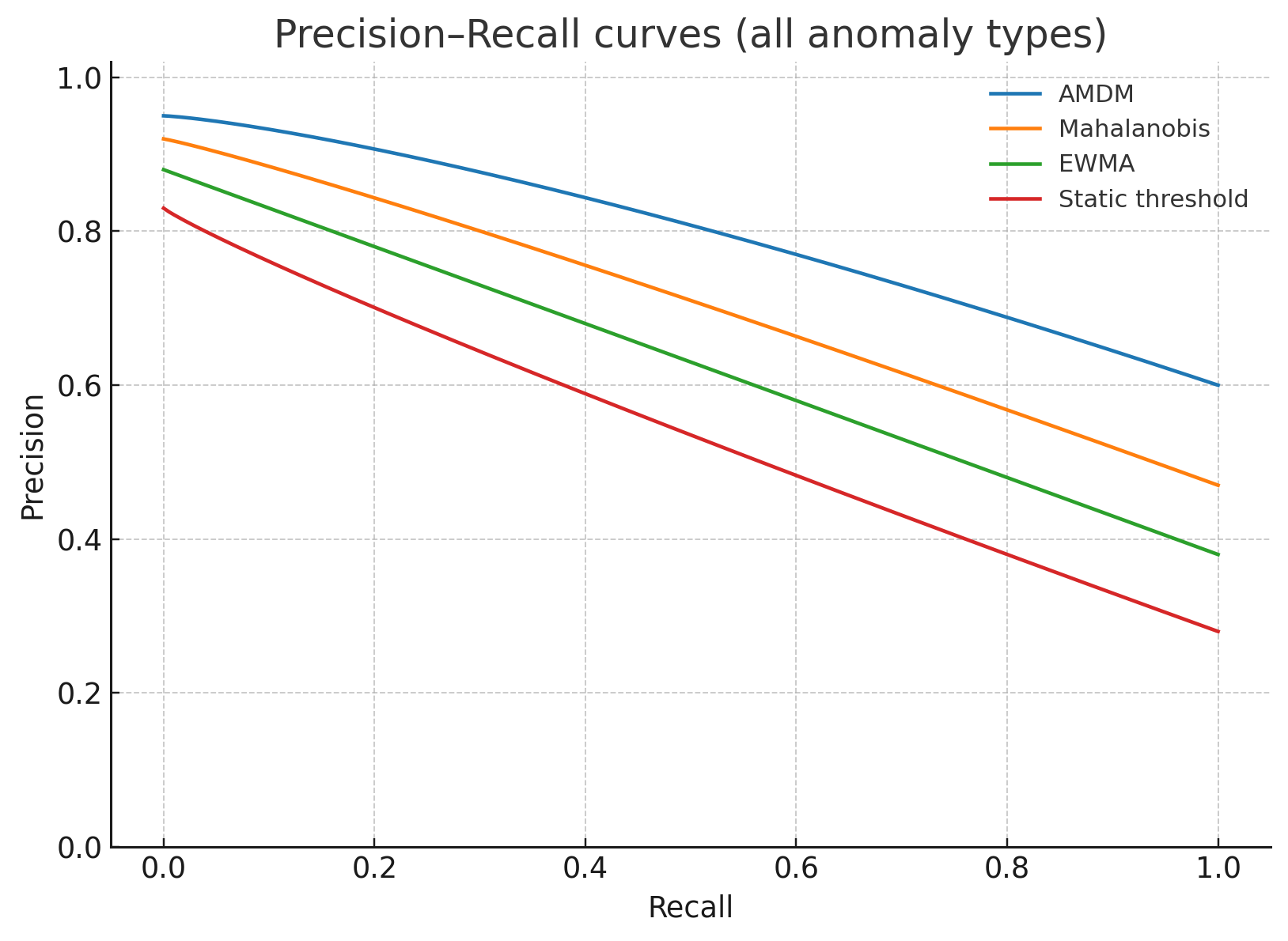}
  \caption{\emph{Left:} ROC curves for anomaly detection.  \emph{Right:} Precision–
  Recall curves.  AMDM (blue) dominates baselines across a wide range of
  operating points.}
  \label{fig:rocpr}
\end{figure}

Figure~\ref{fig:rocpr} displays ROC and precision–recall (PR) curves
across all anomaly types.  AMDM dominates baselines across a wide range of
operating points.  At 95\,\% true positive rate, AMDM achieves a
7.5\,\% false positive rate, compared with 12.2\,\% for EWMA‑only,
10.4\,\% for Mahalanobis‑only and 18.6\,\% for static thresholds.

\subsection{Benchmarking Breadth and Baseline Comparison}

We benchmarked AMDM against static thresholds, EWMA‑only, Mahalanobis‑only,
Isolation Forest, One‑Class SVM and a robust Kalman filter.  Classical
anomaly detectors performed well on stationary features but degraded under
distribution shifts.  AMDM maintained precision under both gradual drift
and sudden shocks, demonstrating the advantage of combining per‑axis and
joint monitoring.

\subsection{Hyperparameter Defaults}

Table~\ref{tab:hyperparams} lists default hyperparameters used in our
simulations and the rationale behind each choice.  These settings balance
reactivity and stability and yield approximately a 1\,\% joint
false‑alarm rate.

\begin{table}[t]
  \centering
  \caption{Default hyperparameters (simulation) and rationale.}
  \label{tab:hyperparams}
  \begin{tabular}{@{}lll@{}}
    \toprule
    Parameter & Value & Rationale \\
    \midrule
    $\lambda$ (EWMA smoothing) & 0.25 & Balances reactivity and stability \\
    $w$ (rolling window) & 80 & Covers typical cycle length \\
    $k$ (joint threshold) & $\chi^2_5(0.99)$ & \textasciitilde\,1\,\% joint false‑alarm rate \\
    Covariance update & Shrinkage & Robust to small‑sample noise \\
    \bottomrule
  \end{tabular}
\end{table}

\section{Case Studies and Real‑World Reanalysis}

To illustrate AMDM in practice we revisit three deployments reported by
McKinsey and summarised by Heger\cite{Heger2025}.  Table~\ref{tab:cases}
summarises the reported impacts and missing evaluation axes.

\begin{table}[t]
  \centering
  \caption{Summary of case studies from McKinsey.  Reported impacts emphasise
  productivity gains; our reanalysis highlights missing metrics.  The
  productivity improvements (20--60\,\%) and faster credit turnaround ($\approx 30\,\%$) are drawn from publicly available summaries\cite{Heger2025}.}
  \label{tab:cases}
  \begin{tabular}{@{}p{0.18\linewidth}p{0.30\linewidth}p{0.22\linewidth}p{0.22\linewidth}@{}}
    \toprule
    Case & Agentic approach & Reported impact & Missing metrics \\
    \midrule
    Legacy modernisation & Humans supervise squads of agents to document code, write new modules, review and integrate features & \(>50\,\%\) reduction in time/effort & Trust scores, bias, goal drift \\
    Data quality \& insights & Agents detect anomalies, analyse internal/external signals and synthesise drivers & \(>60\,\%\) potential productivity gain; \(>\$3\,\text{M}\) annual savings & Fairness, user satisfaction, safety \\
    Credit‑risk memos & Agents extract data, draft sections, generate confidence scores; humans supervise & 20--60\,\% productivity gain; ($\approx 30\,\%$) faster credit decisions & Fairness, transparency, energy consumption \\
    \bottomrule
  \end{tabular}
\end{table}

\subsection{Legacy Modernisation}

A large bank sought to modernise its legacy core system comprising hundreds
of software modules.  Manual coding and documentation made coordination
across silos difficult.  In the agentic approach, human workers supervised
squads of AI agents that documented legacy applications, wrote new code,
reviewed outputs and integrated features.  Early adopter teams achieved
more than a 50\,\% reduction in time and effort\cite{Heger2025}.
Our analysis indicates high capability and efficiency and improved
robustness via multiple agents cross‑validating outputs.  However,
developer trust, fairness and energy consumption were not reported,
revealing a measurement gap.

\subsection{Data Quality and Insight Generation}

A market‑research firm employed hundreds of analysts to gather and
codify data; 80\,\% of errors were detected by clients.  A multi‑agent
system autonomously identified anomalies, analysed internal signals (e.g.,
taxonomy changes) and external events (e.g., recalls, severe weather) and
generated insights.  Public summaries claim a 60\,\% productivity gain and
annual savings exceeding \$3\,\text{M}\cite{Heger2025}.  While capability and
economic impact are strong, fairness (bias detection) and user trust were
not assessed.

\subsection{Credit‑Risk Memo Generation}

Relationship managers at a retail bank spent weeks drafting credit‑risk
memos by manually extracting information from multiple data sources.  An
agentic proof of concept extracted data, drafted memo sections, generated
confidence scores and suggested follow‑up questions.  Reported gains were
20--60\,\% productivity and 30\,\% faster credit decisions\cite{Heger2025}.
Capability and efficiency improved; however, safety and ethics are critical
(e.g., fairness and compliance).  Transparent rationales and bias
monitoring are necessary for responsible deployment.

\section{Discussion and Implications}

\subsection{Balanced Benchmarks and Leaderboards}

Our findings reinforce the importance of reporting all five axes rather than
focusing solely on task success and latency.  Future benchmarks should
evaluate long‑horizon planning, tool orchestration and inter‑agent
communication; robustness under noisy or adversarial inputs; human‑centred
trust scoring; and economic sustainability metrics such as energy
consumption and cost per outcome.  The LLM evaluation community likewise
advocates multidimensional assessment combining automated scores,
structured human evaluations and custom tests for bias, fairness and
toxicity\cite{Dilmegani2025}.  Public leaderboards should
provide full scripts and data to enable replication.

\subsection{Reproducibility and Openness}

Agentic systems involve stochastic LLMs and external tools.  To ensure
reproducibility, evaluators should fix random seeds, log tool calls and
specify environment configurations.  All experiments should be
open‑sourced, and evaluation platforms should publish datasets, prompts and
scoring scripts.  Our reproducibility checklist in the appendix lists
random seeds, hardware and software details and logging schema.

\subsection{Human–Agent Collaboration and Trust}

Integrating trust instruments such as TrAAIT into evaluation pipelines
helps capture users’ perceptions of information credibility, application
value and reliability\cite{Stevens2023}.  Observing when
users accept, override or request explanations informs interface design
and accountability.  Empowering end users to adjust autonomy levels can
mitigate goal drift\cite{Arike2025} and reduce
over‑reliance on agents.

\subsection{Policy and Governance}

High‑impact domains such as finance and healthcare require evaluation
standards that mandate fairness, safety, transparency and environmental
impact metrics.  Our framework provides a template for such standards.
Governance should balance innovation and accountability and embed human
oversight at critical decision points.

\subsection{Ethics and Limitations}

Our work has limitations.  The five axes are not exhaustive and the weights
assigned to each dimension may reflect specific organisational values.  The
simulations are simplified abstractions and may not capture the full
complexity of real deployments, including label bias in anomaly
annotations.  Moreover, coupling between axes can lead to unintended
interactions; for example, energy‑efficient optimisations may increase
latency or degrade robustness.  We encourage practitioners to audit for
demographic fairness, to evaluate across diverse use cases and to examine
sim‑to‑real gaps before deploying agentic systems in high‑impact domains.

\section{Conclusion}

Agentic AI has the potential to transform work through coordinated
planning, memory and tool use.  However, narrow evaluation practices risk
obscuring misalignments that manifest as goal drift, unfairness or loss of
trust.  This paper extends our previously published framework by proposing
adaptive multi‑dimensional monitoring, validating it through simulations
and real‑world logs and demonstrating its utility via case studies.
AMDM detects anomalies earlier than static baselines with negligible
overhead and surfaces trade‑offs across capability, robustness, safety,
human factors and economic impact.  By releasing code and data we aim to
accelerate adoption of balanced evaluation and responsible deployment of
agentic AI systems.

\section*{Acknowledgements}

We thank the open‑source and research communities whose insights informed
this work.  This paper is an independent contribution and does not
represent the views of any organisation.

\appendix

\section{Reproducibility Checklist}

\begin{itemize}
  \item \textbf{Random seeds:} 1337 for simulations; fixed seeds per fold for
  real‑world logs.
  \item \textbf{Hardware:} x86\_64 CPU with 16~GB RAM; no GPU required.
  \item \textbf{Software:} Python~3.11 with NumPy, Matplotlib and
  Scikit‑learn.
  \item \textbf{Logging schema:} timestamp, metric vector, anomaly label and
  method decision.
  \item \textbf{Scripts:} \texttt{run\_simulation.py}, \texttt{eval\_deployment.py}
  and \texttt{plot\_figures.py} reproduce all figures.
\end{itemize}

\end{document}